\documentclass[twocolumn]{cinc}

\usepackage{graphicx}
\usepackage{booktabs}
\usepackage{lipsum}
\usepackage{multirow}
\usepackage{amsmath}
\usepackage{hyperref}
\usepackage{cleveref}
\usepackage{standalone}  
\usepackage{amssymb}
\usepackage{tabularx}
\usepackage{pgf}
\usepackage[justification=justified,singlelinecheck=false]{caption}
\usepackage{tikz}
\usepackage{siunitx}
\usetikzlibrary{positioning, arrows.meta, shapes, backgrounds}
\usetikzlibrary{fit}
\usetikzlibrary{backgrounds}
\usetikzlibrary{calc}

\begin{document}
\bibliographystyle{cinc}

\title{Biomarker-Based Pretraining for Chagas Disease Screening in Electrocardiograms}

\author {Elias Stenhede$^{1,2}$, Arian  Ranjbar$^{1}$ \\
\ \\ 
 $^1$ Medical Technology \& E-health, Akershus University Hospital, Lørenskog, Norway\\
 $^2$ Faculty of Medicine, University of Oslo, Oslo, Norway}

\maketitle

\renewcommand{\heavyrulewidth}{0.08em}
\renewcommand{\lightrulewidth}{0.05em}
\renewcommand{\cmidrulewidth}{0.03em}
\renewcommand{\aboverulesep}{0.4ex}
\renewcommand{\belowrulesep}{0.65ex}
\renewcommand{\abovetopsep}{0.8ex}
\renewcommand{\belowbottomsep}{0.8ex}

\begin{abstract}
    Chagas disease screening via ECGs is limited by scarce and noisy labels in existing datasets. We propose a biomarker-based pretraining approach, where an ECG feature extractor is first trained to predict percentile-binned blood biomarkers from the MIMIC-IV-ECG dataset. The pretrained model is then fine-tuned on Brazilian datasets for Chagas detection. Our 5-model ensemble, developed by the Ahus AIM team, achieved a challenge score of 0.269 on the hidden test set, ranking 5th in Detection of Chagas Disease from the ECG: The George B. Moody PhysioNet Challenge 2025. Source code and the model are shared on GitHub: \href{https://github.com/Ahus-AIM/physionet-challenge-2025}{github.com/Ahus-AIM/physionet-challenge-2025}.
\end{abstract}

\section{Introduction}
    This work was developed as part of the George B. Moody PhysioNet Challenge 2025, which aims to advance automated ECG-based screening methods for Chagas disease~\cite{goldberger_physiobank_2000, 2025ChallengeCinC, 2025ChallengePreprint}. The disease, caused by the parasite Trypanosoma cruzi, remains a significant health burden in Central and South America~\cite{cucunuba_epidemiology_2024}. If left untreated, the infection can result in potentially life-threatening complications. While treatments exist to slow cardiovascular damage, serological testing is inaccessible for much of the at-risk population. Thus, using AI-interpreted ECGs is a promising, resource-efficient option for large-scale screening. Openly accessible ECG databases have been recorded in regions affected by Chagas disease~\cite{cardoso_longitudinal_2016, ribeiro_automatic_2020}, but annotation validity is limited, reducing the effectiveness of traditional supervised deep learning approaches.
    
\section{Methods}
    The following sections describe data sources, preprocessing steps, biomarker-based pretraining strategy, subsequent fine-tuning, and the model architecture used for the challenge.
    \subsection{Data sources}
        The MIMIC-IV-ECG~\cite{gow_mimic-iv-ecg_nodate} is collected across Beth Israel Deaconess Medical Center, Boston, Massachusetts, USA, spanning the period 2008 to 2019. The ECGs in this dataset can be connected to biomarkers by connecting them with the MIMIC-IV dataset~\cite{johnson_mimic-iv_nodate}. Biomarker usage in clinical practice is highly skewed, and to limit the number of tests, selection was based on both prevalence and clinical relevance. The tests used in pretraining are presented in \Cref{tab:selected_tests}. The datasets used for fine-tuning are both collected in Brazil, where Chagas disease is endemic. The CODE15\% dataset is collected by Telehealth Network of Minas Gerais in the period 2010 to 2016, and is paired with self-reported Chagas labels. In contrast, the SaMi-Trop dataset only contains patients with chronic Chagas cardiomyopathy. All utilized datasets are presented in \Cref{tab:datasets}.

        \begin{table}[htbp]
          \centering
          \caption{Biomarkers included for model pretraining (alphabetical).}
          \label{tab:selected_tests}
          \begin{tabularx}{\columnwidth}{@{}l>{\raggedright\arraybackslash}p{0.5\columnwidth}@{}}
            \toprule
            Biomarker\hspace{2.5cm} & Clinical domain \\
            \midrule
            Albumin        & Liver function \\
            Calcium, Total & Electrolytes \\
            Creatinine     & Renal \\
            Hematocrit     & Hematology \\
            Hemoglobin     & Hematology \\
            INR(PT)        & Coagulation \\
            NTproBNP       & Cardiac \\
            Potassium      & Electrolytes \\
            Red Blood Cells& Hematology \\
            Troponin T     & Cardiac \\
            Urea Nitrogen  & Renal \\
            \bottomrule
           \end{tabularx}
        \end{table}
 
        \begin{table}[htbp]
          \centering
          \caption{Datasets used for model development. In pretraining, not all ECGs were included, as each ECG had to occur within 24 hours of at least one included test.}
          \label{tab:datasets}
          \begin{tabularx}{\columnwidth}{@{}llll@{}}
            \toprule
            Dataset        & \# ECGs  & \# Patients & Step           \\
            \midrule
            MIMIC-IV-ECG & 523,275  & 102,511     & Pretraining  \\
            CODE15\%     & 345,779  & 233,770     & Fine-tuning\\
            SaMi-Trop    & 1,959    & 1,959       & Fine-tuning\\
            \bottomrule
          \end{tabularx}
        \end{table}
    
    \subsection{Preprocessing steps}
        Preprocessing involves standardizing ECG signals, resolving label inconsistencies, and normalizing biomarker values for pretraining. Details for each component are provided in the subsections below.
        \subsubsection{Electrocardiograms}
        Across all datasets, ECGs are resampled to 400 Hz, followed by dropping leads I, II, III, and aVR, as they are linear combinations of aVL and aVF defined by Einthoven's Law. During training, a single two-second snippet is randomly extracted from each ECG. All snippets are normalized to have zero mean and unit variance before being fed into the model.
    
        \subsubsection{Chagas labels}
        In the CODE15\% dataset, a total of 1,825 patients have ECGs labelled both as Chagas-positive and negative. In an attempt to reduce label noise, all ECGs for these patients are labelled using the average positive proportion of ECGs. The aim is to reduce noise, without discarding these patients, to preserve sample size while minimizing label uncertainty.
    
        \subsubsection{Biomarkers}
        Normalizing biomarker values is challenging due to differing distributions and the need to preserve clinically relevant cut-offs. We address this by replacing each value with its percentile rank (binned into 100 intervals). Biomarker results are matched to ECGs by timestamp; ECGs without a paired result within 24 hours are excluded from pretraining.
    
    \subsection{Training strategy}
        All model training is performed on the setup detailed in \Cref{tab:env}. Across all model training, the Muon optimizer is used for parameters with dimension $\geq$2, as it has proved to converge faster and yield better results~\cite{noauthor_muon_nodate}. For the remaining parameters, i.e., biases and parameters in normalization layers, Adam is used~\cite{kingma_adam_2017}. The same learning rate is used for both optimizers. For clarity, the pretraining and fine-tuning training recipes are described one at a time.
        
        \begin{table}[!htbp]
            \caption{Development environments and hardware used in model training.}
            \label{tab:env}
            \centering
            \begin{tabularx}{\columnwidth}{@{}ll@{}}
                \toprule
                Component                   & Specification \\
                \midrule
                System                      & Debian 12 \\
                CPU                         & Intel Core i9-14900KF \\
                RAM                         & 2$\times$48\,GB; 4800\,MT/s \\
                GPU                         & NVIDIA GeForce RTX 5090 \\
                CUDA version                & 12.9 \\
                Programming language        & Python 3.12 \\
                Deep learning framework     & PyTorch 2.7.1+cu128\\
                \bottomrule
            \end{tabularx}
        \end{table}
        
        \subsubsection{Biomarker based pretraining}
            \label{sec:pretraining}
            The pretraining task is formulated as a classification problem for each biomarker, where the target is the percentile rank of the test result. Most ECGs are not paired with all selected biomarkers; in these cases, the loss is not computed for those missing biomarker-ECG pairs. Using classification instead of regression enables the model to predict a probability distribution over possible blood test values, which provides richer supervision and reduces bias toward the mean value of each biomarker. Specifically, the model outputs logits with shape $\mathbb{R}^{\mathrm{batch\_size} \times 100 \times \mathrm{T}}$ with $\mathrm{T}$ denoting the number of tests. Cross-entropy loss is computed over the 100 percentile bins for each available biomarker. The parameters used in pretraining are listed in \Cref{tab:pretraining}.
            
            \begin{table}[!htbp]
                \caption{Training parameters for biomarker-based pretraining.}
                \label{tab:pretraining}
                \centering
                \begin{tabularx}{\columnwidth}{@{}ll@{}}
                    \toprule
                    Parameter\hspace{2.5cm}  & Value \\
                    \midrule
                    Batch size               & 64 \\
                    Optimizer                & Muon and Adam\\
                    Muon momentum            & 0.95 \\
                    Learning rate            & 0.0037 \\
                    Loss function            & Cross-entropy\\
                    \bottomrule
                \end{tabularx}
            \end{table}
    
            To mitigate the label sparsity introduced by percentile binning, we use a decoupled regularization step applied after each optimizer step. In essence, we enforce the inductive bias of neighbouring bins sharing directions in weight space.
            We treat the final fully‐connected layer’s weight as  
            $$
            W\;\in\;\mathbb{R}^{\mathrm{(100\times T)\times F}}
            $$
            where \(F\) is the feature dimension of the activations just before the last layer and \(T\) the number of tests.  After the usual gradient-based update, we apply an in‑place smoothing update with  
            $
            \alpha = \eta\,\beta,
            $
            where $\eta$ is the learning rate and $\beta$ a fixed bin‑smoothing factor. For each weight row $W_i$, we then perform
            \begin{align*}
            W_i \;\leftarrow\;
            \begin{cases}
            (1-\tfrac{\alpha}{2})W_1 + \tfrac{\alpha}{2}W_2, & i=1,\\
            (1-\alpha)W_i + \tfrac{\alpha}{2}(W_{i-1}+W_{i+1}), & 2\le i\le 99,\\
            (1-\tfrac{\alpha}{2})W_{100} + \tfrac{\alpha}{2}W_{99}, & i=100.
            \end{cases}
            \end{align*}
            Because this smoothing runs after back‑propagation and does not contribute to any gradients, it adds negligible overhead. The bin-smoothing factor $\beta$ is set to 1 in this work, and it can be optimized as a hyperparameter.
        
        \subsubsection{Fine-tuning for Chagas screening}
            The fine-tuning step starts by initializing a feature extractor with the weights obtained in the pretraining step, and dropping the last linear layer mapping to biomarkers, instead replacing it with a new randomly initialized layer that will be trained to output logits corresponding to the probability of Chagas disease. Although the CODE15\% and SaMi-Trop datasets contain both self-reported and strong labels, they are pooled. To ensure good generalization, 5-fold cross-validation is used to train 5 models, with each model being selected at the epoch where the validation loss is lowest. Parameters used in this step are detailed in \Cref{tab:finetuning}.
            
            \begin{table}[!htbp]
                \caption{Parameters used during fine-tuning.}
                \label{tab:finetuning}
                \centering
                \begin{tabularx}{\columnwidth}{@{}ll@{}}
                    \toprule
                    Parameter\hspace{2.5cm}  & Value \\
                    \midrule
                    Batch size               & 128 \\
                    Optimizer                & Muon and Adam\\
                    Muon momentum            & 0.95 \\
                    Learning rate            & 0.001 \\
                    Loss function            & Binary cross-entropy\\
                    \bottomrule
                \end{tabularx}
            \end{table}
    
    \subsection{Model architecture}
        The model is based on the InceptionTime architecture~\cite{fawaz_inceptiontime_2020}, with minor modifications. Before the inception blocks, we include two convolutional layers followed by batch normalization and GELU activation. Each convolution uses a kernel size of 5 and a stride of 2, which ensures that the receptive field within the network is sufficiently large. The parameters for the InceptionTime network are summarized in \Cref{tab:inception}.
        
        \begin{table}[!htbp]
            \caption{Hyperparameters for the InceptionTime network.}
            \label{tab:inception}
            \centering
            \begin{tabularx}{\columnwidth}{@{}ll@{}}
                \toprule
                Parameter\hspace{2.5cm}  & Value \\
                \midrule
                Number of blocks         & 6 \\
                Kernel sizes             & 9, 19, 39 \\
                Number of filters        & 32 \\
                Bottleneck channels      & 32 \\
                \bottomrule
            \end{tabularx}
        \end{table}
    \subsection{Final ensemble}
        During inference, each ECG is divided into ten overlapping two-second segments, which are individually standardized and processed by the ensemble models. The logits produced by each model are transformed with a sigmoid function, and the resulting probabilities are averaged across all segments and ensemble members. 

\section{Results}
\begin{figure*}[ht]
        \centering
        \resizebox{\textwidth}{!}{\input{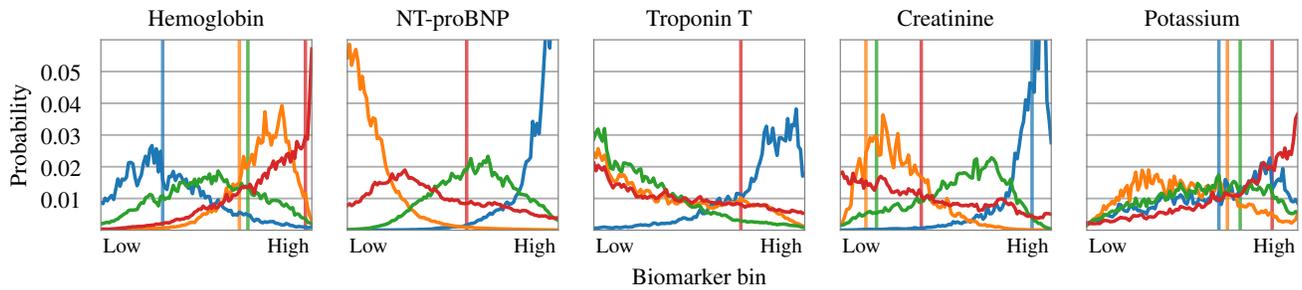}}
        \caption{Shows the predicted probability distributions over blood test percentiles for four ECGs taken from four different patients (one colour per patient). The vertical lines represent the actual values for the same patients, measured within 24 hours of ECG recording. The selected biomarkers represent a subset of the biomarkers used in pretraining.}
        \label{fig:percentiles}
    \end{figure*}

    The validation loss for biomarker-based pretraining reached its minimum after three epochs. When ranking biomarkers by validation-set perplexity, the model predicted NT-proBNP most accurately, followed by albumin, hemoglobin, and troponin. Predicted probability distributions for selected biomarkers are illustrated in \Cref{fig:percentiles}.

    The fine-tuned model was evaluated using the official challenge metric, defined as the number of Chagas-positive cases in the top 5\% of ECGs sorted by model-predicted risk, divided by the total number of cases. When evaluating the model using 5-fold cross-validation on the development set, a mean challenge score of 0.439 (SD = 0.010) and an AUC-ROC of 0.840 (SD = 0.008) were achieved. When deploying the ensemble on the hidden validation set, it achieved a challenge score of 0.412, and on the hidden test set, a challenge score of 0.269, resulting in a ranking of 5 out of the 40 teams eligible for ranking, also displayed in \Cref{tab:results}. Notably, larger models resulted in better cross-validated scores on the development set, but not on the hidden validation set.

    \begin{table}[!htbp]
        \caption{Challenge score on the hidden test set.}
        \label{tab:results}
        \centering
        \begin{tabularx}{\columnwidth}{@{}lll@{}}
            \toprule
            Team\hspace{1.7cm} & Challenge score\hspace{0.7cm}   & Rank \\
            \midrule
            Ahus AIM & 0.269 & 5 \\
            \bottomrule
        \end{tabularx}
    \end{table}

\section{Discussion and conclusions}
    While traditional approaches to label noise include partial self-supervision, label correction techniques, early stopping, or robust loss functions~\cite{song_learning_2022}, our method leverages clinical biomarkers to guide the model in learning physiologically relevant ECG features before fine-tuning for Chagas disease detection. One notable aspect of our study is the geographical and clinical diversity of the datasets. Pretraining was conducted on datasets collected in the USA, where Chagas disease is rare, whereas fine-tuning was performed on Brazilian datasets where the disease is endemic. This split underscores the importance of evaluating domain shift and raises questions regarding the generalizability of features learned from one population to another. The degree to which biomarker-driven pretraining can yield transferable and robust ECG representations across diverse settings remains an area for further study.


    Using the ECG to predict blood samples beyond the classification of abnormal values or regression has not been rigorously explored. It may be valuable in itself, and not only as a pretraining task. By letting the model output probability distributions over test results, it can be adapted to different cut-off values without retraining.

\section*{Acknowledgements}
We thank Akershus University Hospital for the funding that made this work possible.

\bibliography{refs}

\begin{correspondence}
Elias Stenhede\\
Sykehusveien 25, 1478 Nordbyhagen, Norway\\
elias.stenhede@ahus.no
\end{correspondence}

\end{document}